\documentclass[a4paper,num-refs]{oup-contemporary}
\usepackage{graphicx}
\usepackage{amsmath}
\usepackage{amssymb}
\usepackage{booktabs}
\usepackage[utf8]{inputenc}
\usepackage{siunitx}
\usepackage{tikz}
\usepackage{multicol}
\usetikzlibrary{shapes.geometric, arrows, positioning, calc}

\usepackage{xcolor} % 用于文字标红

\newif\ifwendireviewmode

\wendireviewmodefalse  % <--- 这一行存在时（或者上面被注释）：实名模式 + 正常字

\newif\ifttreviewmode

\ttreviewmodefalse  % <--- 这一行存在时（或者上面被注释）：实名模式 + 正常字

\newcommand{\difftext}[1]{%
  \ifwendireviewmode%
    \textcolor{red}{#1}% 如果开关打开，显示红色
  \else%
    #1% 如果开关关闭，正常显示（黑色）
  \fi%
}

\newcommand{\difftextbytt}[1]{%
  \ifttreviewmode%
    \textcolor{red}{#1}% 如果开关打开，显示红色
  \else%
    #1% 如果开关关闭，正常显示（黑色）
  \fi%
}

\title{Diagnostic Performance of Universal-Learning Ultrasound AI Across Multiple Organs and Tasks: the UUSIC25 Challenge}

\ifwendireviewmode
    \author{* \\ *}
\else
    \author[1]{Zehui Lin}
    \author[2]{Luyi Han}
    \author[2]{Xin Wang}
    \author[1,3]{Ying Zhou}
    \author[1,3]{Yanming Zhang}
    \author[2]{Tianyu Zhang}
    \author[4]{Lingyun Bao}
    \author[1]{Jiarui Zhou}
    \author[1]{Yue Sun}
    \author[6]{Jieyun Bai}
    \author[7]{Shuo Li}
    \author[5]{Shandong Wu}
    \author[8]{Dong Ni}
    \author[2]{Ritse Mann}
    \author[10]{Wendie Berg}
    \author[3]{Dong Xu\authfn{1}}
    \author[1]{Tao Tan\authfn{1}}
    \author[ ]{the UUSIC25 Challenge Consortium\authfn{2}}

    \affil[1]{Macao Polytechnic University, Macao, China}
    \affil[2]{Netherlands Cancer Institute, Amsterdam, The Netherlands}
    \affil[3]{Zhejiang Cancer Hospital, Hangzhou, China}
    \affil[4]{The First People’s Hospital of Hangzhou, Affiliated Hangzhou Hospital of Nanjing Medical University, Hangzhou, China}
    \affil[5]{Department of Radiology, University of Pittsburgh,Pittsburgh, PA, USA}
    \affil[6]{College of Information Science and Technology, Jinan University, Guangzhou, China}
    \affil[7]{Departments of Biomedical Engineering, and Computer and Data Science, Case Western Reserve University, USA}
    \affil[8]{School of Biomedical Engineering, Shenzhen University, Shenzhen, China}
    \affil[10]{University of Pittsburgh Medical Center, Pittsburgh, PA, USA}
\fi

\authnote{\authfn{1}Corresponding author: xudong@zjcc.org.cn, taotan@mpu.edu.mo}
\authnote{\authfn{2}A complete list of the UUSIC25 Challenge Consortium members and their affiliations is provided in the Appendix.}

\papercat{Paper}

\runningauthor{Lin et al.}

\jvolume{00}
\jnumber{0}
\jyear{2025}

\begin{document}

\begin{frontmatter}
\maketitle

% ===================================================================
% SECTION: ABSTRACT (负责人: 泽辉)
% ===================================================================
\begin{abstract}

    \textbf{IMPORTANCE} Modern ultrasound systems are universal, economically cost-effective, and portable diagnostic tools capable of imaging the entire body. However, current artificial intelligence (AI) solutions remain fragmented into single-task, organ-specific tools. This critical \textbf{gap} between hardware versatility and software specificity limits workflow integration and the clinical utility of AI in general ultrasonography.

    \textbf{OBJECTIVE} To evaluate the diagnostic accuracy, versatility, and computational efficiency of single general-purpose deep learning models in performing multi-organ classification and segmentation tasks compared across diverse anatomical regions.

    \textbf{DESIGN, SETTING, AND PARTICIPANTS} The Universal UltraSound Image Challenge 2025 (UUSIC25), an international competition was held from July to September 2025. Participants developed algorithms using a training set of 11,644 images aggregated from 12 discrete sources (9 public datasets and 3 partner hospitals). Algorithms were evaluated on a fully independent, multi-center private test set of 2,479 images, composed of held-out internal samples and a cohort from an external center completely unseen during training to assess generalization.

    \textbf{MAIN OUTCOMES AND MEASURES} Primary outcomes were diagnostic performance (Dice Similarity Coefficient [DSC] for segmentation; Area Under the Receiver Operating Characteristic Curve [AUC] and Accuracy for classification) and computational efficiency (inference time and GPU memory). Performance metrics are reported as bootstrap means with 95\% confidence intervals (CIs).

    \textbf{RESULTS} A total of 15 valid algorithms were evaluated. The top-ranking algorithm (SMART) achieved a macro-averaged DSC of 0.854 (95\% CI, 0.848-0.861) across 5 segmentation tasks and a macro-averaged AUC of 0.766 (95\% CI, 0.718-0.815) for binary classification tasks. While models demonstrated high capability in anatomical segmentation (e.g., fetal head DSC: 0.942 [95\% CI, 0.934-0.948]), performance variability was observed in complex diagnostic tasks subject to domain shift. Specifically, in breast cancer molecular subtyping, the top model yielded a limited baseline AUC of 0.571 (95\% CI, 0.505-0.643) on the internal domain, which further dropped to 0.508 (95\% CI, 0.469-0.544) on the unseen external center, highlighting both the inherent difficulty of this diagnostic task and the challenge of generalization.

    \textbf{CONCLUSIONS AND RELEVANCE} General-purpose AI models can achieve high diagnostic accuracy and efficiency across multiple ultrasound tasks using a single network architecture. However, significant performance degradation on data from unseen institutions suggests that future development must prioritize domain generalization techniques before clinical deployment is feasible.

\end{abstract}

\end{frontmatter}

%%% Key points will be printed at top of second page
\begin{keypoints*}
\begin{itemize}
\item \textbf{Question} How effectively can a single general-purpose deep learning model handle multi-organ segmentation and classification tasks in clinical ultrasound?
\vspace{1em}
\item \textbf{Findings} In the UUSIC25 challenge involving 15 algorithms and 16,021 images across 7 anatomical regions, the winning query-driven Transformer model achieved high diagnostic accuracy (e.g., 0.942 Dice for fetal head segmentation, 0.837 AUC for breast malignancy) and efficiency. Notably, these unified models demonstrated robust generalization on a fully private, multi-center test set containing data from a completely unseen institution.
\vspace{1em}
\item \textbf{Meaning} These results suggest that developing high-performing, "all-in-one" clinical ultrasound AI systems is feasible, moving beyond the fragmented single-task paradigm; this paves the way for next-generation AI assistants that can streamline workflows and adapt to diverse clinical scenarios without manual intervention.
\end{itemize}
\end{keypoints*}

% ===================================================================
% SECTION: INTRODUCTION (负责人: 天宇)
% ===================================================================
\section{Introduction}

% [Revised Paragraph 1: Establishing the "Hardware-Software Gap"]
\difftextbytt{In clinical practice, a single modern ultrasound system, equipped with a versatile family of probes, serves as a "universal" diagnostic tool capable of imaging the entire human body—from the thyroid and breast to the liver and fetus \cite{van2024flexible,kaselitz2025point,arnaout2021ensemble,houssami2025breast,conner2025perspectives,baloescu2025artificial}. However, the current landscape of artificial intelligence (AI) in ultrasonography stands in stark contrast to this hardware versatility. Existing AI solutions are predominantly fragmented, developed as "single-organ, single-task" tools \cite{cid2021effect,maganti2025cardiopulmonary,nafade2024value, gao2024explainable, zhang2023predicting}. This discrepancy creates a fundamental \textbf{utility gap}: while the sonographer seamlessly switches probes to scan different organs within a single session \cite{wang2020anatomy,frija2021improve,vega2025overcoming}, the supporting AI software cannot.}

% [Revised Paragraph 2: The Motivation]
\difftextbytt{This "one model, one problem" paradigm imposes severe limitations on clinical integration \cite{fiorentino2023review}. It requires hospitals to deploy, validate, and maintain dozens of isolated algorithms, leading to prohibitive computational costs and complex user interactions \cite{peng2021deep,qian2021prospective,slimani2023fetal,tong2023integration,survarachakan2022deep,yan2024domain}. To bridge this gap, we posed a fundamental question: \textit{Can a single, general-purpose deep learning model replicate the versatility of a sonographer?} To answer this, we organized the Universal UltraSound Image Challenge 2025 (UUSIC25) inviting the global community to develop unified architectures capable of performing segmentation and classification across seven anatomical regions simultaneously.}

\difftext{In parallel, there is growing interest in general purpose or foundation models for ultrasound that learn shared representations across organs and tumors \cite{jiao2024usfm,jiang2025pretraining,lin2025orchestration,chen2024multi}. Many malignant lesions, regardless of anatomical site, exhibit overlapping sonographic patterns, including irregular margins, heterogeneous echotexture, and abnormal posterior acoustic behavior, whereas normal organs show characteristic and reproducible structural motifs \cite{peng2021deep,chen2023thyroid,pawlowska2024curated}. A model trained jointly on several organs may capture these common features and differences, potentially improving generalization and offering more interpretable decision patterns. Clarifying which groups of organs benefit most from cross organ learning, and how organ level structure and tumor level appearance jointly contribute to performance and interpretability, is essential for the rational design of future general purpose ultrasound AI systems \cite{jiao2024usfm,jiang2025pretraining,lin2025orchestration,cerrolaza2019computational}.}

To address these gaps, the Universal UltraSound Image Challenge 2025 (UUSIC25) was organized as an international competition, held in conjunction with MICCAI 2025, focusing on general purpose multi organ ultrasound AI for both classification and segmentation tasks. \difftext{Participants were asked to develop single models that jointly handled seven different organs and clinical scenarios, including organ level segmentation and clinically critical classification problems such as breast cancer malignancy assessment, using training data that combined publicly available datasets with additional curated cohorts.} The challenge design introduced three key innovations aimed at overcoming the current fragmented landscape: (1) a combined ranking based on both segmentation and classification performance, in order to promote genuine multi task capability in general purpose models; (2) explicit treatment of computational efficiency as a core ranking dimension, to encourage algorithms that are not only accurate but also feasible for deployment at the point of care; and (3) final evaluation on a fully private, multi-center test set. Distinct from traditional zero-shot benchmarks, we strategically incorporated a stratified minority (30\%) of the private cohort into the training phase. This design was intended to assess data-efficient domain adaptation and to test whether general-purpose models can leverage limited local samples to mitigate domain shift. This evaluation protocol closely approximates the calibration-before-deployment workflow typically required when introducing AI systems into new clinical centers.

In this study, we report the design, conduct, and outcomes of the UUSIC25 challenge, “Diagnostic Performance of General-Purpose Ultrasound AI Across Multiple Organs and Tasks.” We describe the dataset composition, task definitions, and evaluation framework, and we systematically analyze the submitted algorithms, with particular attention to general purpose foundation architectures and parameter efficient designs. By comparing diagnostic performance, robustness to domain shift, and computational efficiency across diverse multi-task models in a standardized, multi-center setting, this work provides a community-based assessment of the current state of general-purpose ultrasound AI.

To the best of our knowledge, UUSIC25 is the first international challenge specifically dedicated to ultrasound foundation models, in which single general-purpose models are required to jointly perform multi-organ segmentation and classification on a fully private, multi-center ultrasound test set while explicitly accounting for computational efficiency. This benchmark is intended to guide the development of next-generation “all-in-one” clinical ultrasound AI systems and to inform the design of future prospective impact studies.

% ===================================================================
% SECTION: METHODS (负责人: 路易)
% ===================================================================
\section{Methods}

\subsection{Challenge Design and Data}

To benchmark the capability of single models in handling multi-organ and multi-task scenarios, we curated a large-scale, heterogeneous ultrasound dataset comprising \textbf{16,021 images} from diverse global sources. The data accrual, exclusions, and partitioning strategies are detailed in the study flow diagram (\textbf{Figure 1}). The complete pre-specified challenge design, including organization details, and participation rules, is provided in the \textbf{Original Challenge Proposal (eAppendix 1 in Supplement)}.

The cohort aggregates data from 9 public repositories (n=10,010) and internal retrospective collections from 3 partner hospitals (n=6,011), covering seven anatomical regions: breast, thyroid, liver, kidney, fetal head, cardiac, and appendix. The private in-house data were sampled from Zhejiang Cancer Hospital and Hangzhou First People's Hospital in China, and the Netherlands Cancer Institute (NKI). This study was approved by the Medical Ethics Committee of Zhejiang Cancer Hospital (Approval No. IRB-2024-494 (IIT)), Netherlands Cancer Institute (Approval No. IRBd21-058). The requirement for informed consent was waived due to the retrospective nature of the study, and all images were de-identified and converted to a standardized format.

\subsection{Data Partitioning Strategy}
As illustrated in \textbf{Figure 1}, we employed a stratified splitting strategy designed to simulate real-world clinical deployment challenges. 
\textbf{Training Set ($n=11,644$):} To encourage the learning of robust, generalizable features, the training set included 100\% of the public datasets combined with a stratified 30\% subset of the internal private data. This inclusion of a minority fraction of private data was intentional, mimicking a "calibration" phase where a general model is adapted to a local clinical environment using limited samples.
\textbf{Validation and Test Sets:} The remaining 70\% of the internal private data were split equally into validation ($n=1,898$) and test sets. Crucially, to rigorously assess out-of-distribution (OOD) generalization, the dataset from the external partner (NKI, n=512) was held out entirely from the training and validation phases and used exclusively for final testing ($n = 2,479$). This design ensures that the reported performance metrics reflect the model's ability to handle completely unseen devices and patient populations.

\begin{figure*}[t]
\centering
\includegraphics[width=0.95\textwidth]{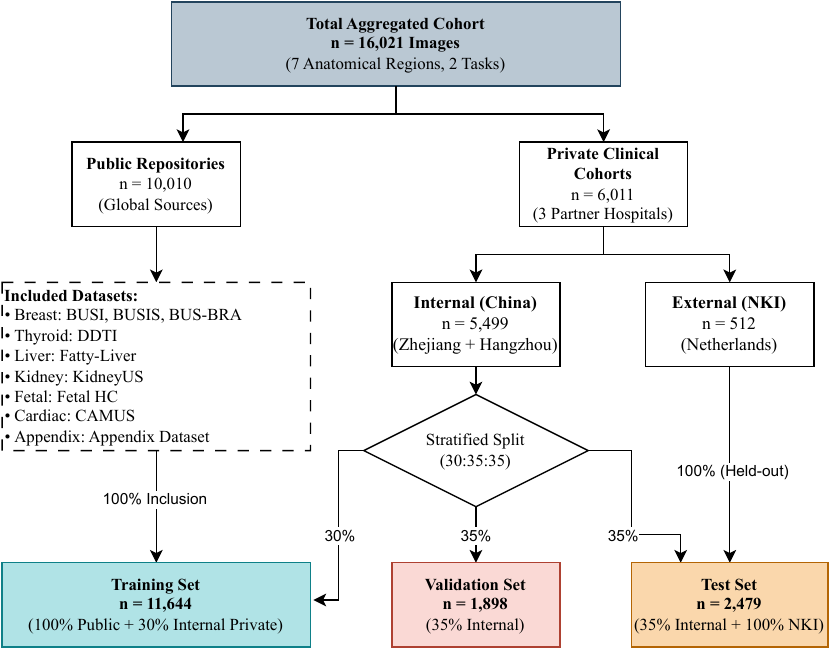} 
\caption{\textbf{Study Flow Diagram.} Data collection, source attribution, and stratification logic. 
The diagram illustrates the explicit separation of data streams: public datasets were utilized exclusively for training to promote generalization (n=10,010), while internal private data were stratified across all sets (n=5,499). Data from the external center (NKI, n=512) served as a strictly held-out test set. 
\textbf{Data Sources:} Public datasets included BUSI \cite{al2020dataset}, BUSIS \cite{zhang2022busis}, and BUS-BRA \cite{gomez2024bus} for breast; DDTI \cite{pedraza2015open} for thyroid; Fatty-Liver \cite{byra2018transfer} for liver; KidneyUS \cite{singla2023open} for kidney; Fetal HC \cite{van2018automated} for fetal head; CAMUS \cite{leclerc2019deep} for cardiac; and the Appendix Dataset \cite{marcinkevics_2023_7711412}. 
\textbf{Abbreviations:} Internal centers in China include \textit{Zhejiang Cancer Hospital} and \textit{Hangzhou First People's Hospital}. The external center in the Netherlands refers to the \textit{Netherlands Cancer Institute (NKI)}.}
\label{fig:consort}
\end{figure*}

\subsection{Evaluation Protocol and Ranking Scheme}

To ensure reproducibility and fairness, all participants were required to submit their algorithms as encapsulated Docker containers via the CodaLab competition platform. This approach prevented any manual intervention during the inference phase on the private validation and test sets.

\textbf{Evaluation Metrics.} We employed a multi-dimensional metric system to assess the algorithms. For segmentation tasks, we utilized the \textbf{Dice Similarity Coefficient (DSC)} to measure volumetric overlap and the \textbf{Normalized Surface Dice (NSD)} to assess boundary adherence at a clinically relevant tolerance. For classification tasks, discriminative performance was evaluated using the \textbf{Area Under the Receiver Operating Characteristic Curve (AUC)} and \textbf{Accuracy (ACC)}. Crucially, to assess clinical deployability, we measured \textbf{Inference Time} (seconds total test cases) and \textbf{Maximum GPU Memory Usage} (GB).

\textbf{Ranking Scheme.} Algorithms were ranked based on a composite score that prioritized diagnostic accuracy (weighted 70\%) while penalizing excessive computational resource consumption (weighted 30\%). This scoring design explicitly discouraged computationally heavy ensemble models to ensure feasibility for real-time clinical deployment. Full ranking details are available in the Supplement.

\subsubsection{Main Outcomes and Measures}

Primary outcomes were diagnostic performance and computational efficiency. Diagnostic accuracy was assessed using Dice Similarity Coefficient (DSC) and Normalized Surface Dice (NSD) for segmentation, and Area Under the Curve (AUC) and Accuracy (ACC) for classification. Specifically, breast tasks included both binary malignancy classification and a fine-grained 4-class molecular subtyping task evaluated on unseen centers to test generalization. Computational efficiency was measured by total inference time and maximum GPU memory usage. Detailed definitions and sensitivity analyses for boundary tolerance are provided in \textbf{eTable 1-4 in Supplement}.

% ===================================================================
% SECTION: RESULTS (负责人: 泽辉, 鑫)
% ===================================================================
\section{Results}

\subsection{Overview of Generalist Capabilities}

\begin{figure*}[!htbp]
\centering
\includegraphics[width=0.85\textwidth]{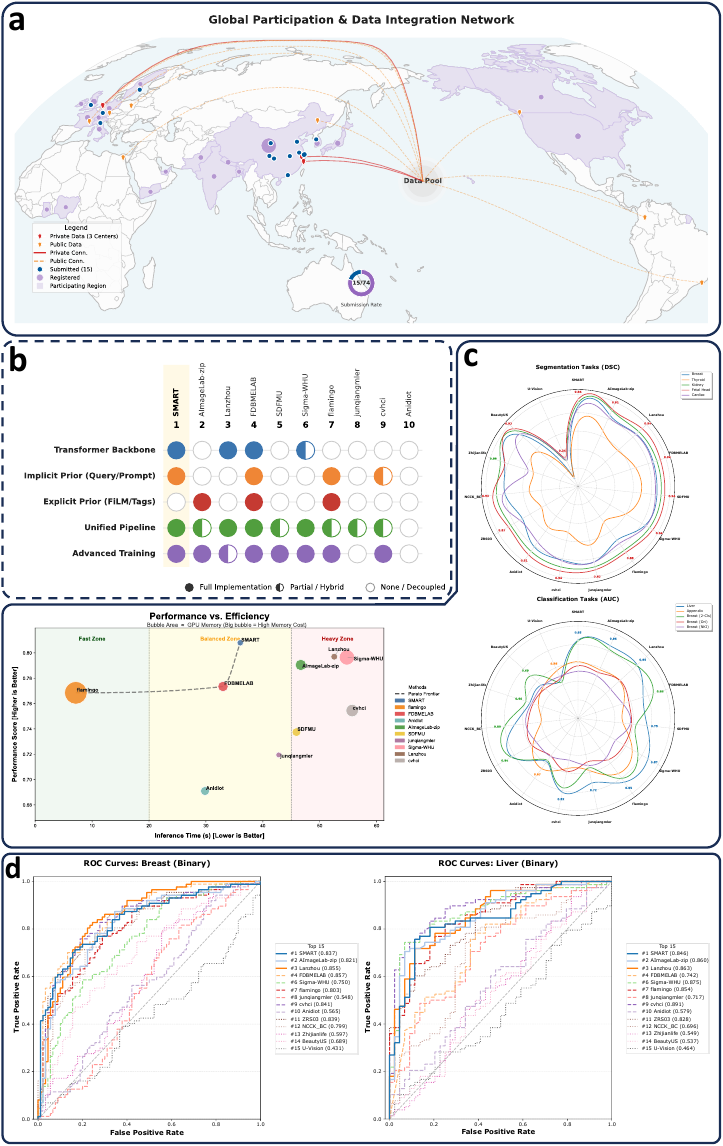} 
\caption{
\textbf{Global Landscape, Methodological Diversity, and Performance Benchmarking of the UUSIC25 Challenge.}
\textbf{(a) Global Participation \& Data Integration Network.} The map visualizes the geographical distribution of participating research teams and the origins of clinical datasets, spanning five continents. This broad geographical coverage was designed to mitigate site-specific biases often found in single-center datasets and highlights the diversity of the patient population used for model training and evaluation.
\textbf{(b) Methodological Configuration Matrix.} A summary of architectural choices for the top-10 teams, categorizing backbones (Transformer vs. CNN), prior integration (implicit queries vs. explicit tags), and training pipelines. \textbf{Note:} This panel provides a high-level overview; detailed technical descriptions and architecture diagrams for all valid algorithms are provided in the \textbf{eMethods} in the Supplement.
\textbf{(c) Multi-Organ Diagnostic Versatility and Efficiency Pane.} Radar charts displaying segmentation (Dice Similarity Coefficient, DSC) and classification (Area Under Curve, AUC) performance across anatomical regions. The winning model (SMART) demonstrates consistent coverage across tasks. The scatter plot from the Efficiency Pane (bubble size = GPU memory) positions algorithms based on inference speed and aggregate performance score, identifying the "Pareto optimal" zone for clinical deployment.
\textbf{(d) Diagnostic Precision (ROC Curves).} Receiver Operating Characteristic curves for the two most distinct clinical tasks: Breast Malignancy (focal lesion) and Fatty Liver (diffuse texture), evaluated on the private test set.
}
\label{fig:composite_results}
\end{figure*}

To comprehensively benchmark the capabilities of the submitted algorithms, we analyzed performance across anatomical versatility, discriminative power, and computational efficiency (Figure \ref{fig:composite_results}).

\difftextbytt{\textbf{Clinical Versatility: Can One Model Handle Diverse Anatomies? (Figure \ref{fig:composite_results}B.}}
The evaluation revealed distinct performance profiles across anatomical regions. Top-performing algorithms maintained high segmentation capability (DSC $> 0.85$) across Breast, Cardiac, Fetal Head, and Kidney datasets. However, performance consistently dipped for the Thyroid task, reflecting the intrinsic difficulty of delineating isoechoic nodules compared to structures with distinct boundaries. In classification tasks, while general-purpose models demonstrated broad applicability, trade-offs were observed; algorithms optimized for breast malignancy screening often showed reduced sensitivity in non-cancerous tasks such as appendicitis diagnosis.

\difftextbytt{\textbf{Deployment Feasibility: High Performance with Minimal Resource Cost (Figure \ref{fig:composite_results}B).}}
Clinical translation requires models that are not only accurate but also deployable. We identified a distinct trade-off between diagnostic accuracy and computational resource consumption. The winning algorithm (SMART) achieved an optimal balance, delivering top-tier diagnostic accuracy with a moderate inference time (36.03s) and a low memory footprint (0.59GB), making it suitable for standard clinical consoles. In contrast, while the fastest algorithm achieved inference speeds of 7.14s, it incurred a massive memory cost (12.41GB), limiting its utility on portable ultrasound devices. Most Transformer-based solutions converged in the center of the efficiency spectrum, suggesting an emerging architectural standard for medical foundation models.

\difftextbytt{\textbf{Diagnostic Precision: Does Generalization Compromise Accuracy? (Figure \ref{fig:composite_results}D).}}
ROC analysis highlighted the robustness of general-purpose models in handling disparate visual tasks. In the focal task of breast malignancy classification, the top models achieved high specificity in the low false-positive rate region, a critical trait for minimizing unnecessary biopsies. Similarly, in texture-based fatty liver diagnosis, these architectures effectively captured diffuse echotexture changes with AUCs exceeding 0.85. However, performance was notably lower for complex diagnostic tasks subject to high inter-observer variability, such as appendicitis and molecular subtyping (see \textbf{eFigure 2 in Supplement}), underscoring current limits in extracting sub-visual biomarkers. Furthermore, visual inspection of segmentation probability maps confirmed that top-performing models maintained high boundary adherence across varying morphological complexities, from large fetal heads to irregular breast tumors (\textbf{eFigure 1 in Supplement}).

\subsection{Challenge Participation and Methods Overview}

The initiative attracted 74 registered teams from Asia, Europe, Africa and America, integrating data sources spanning multiple continents to test cross-population generalization (\textbf{Figure \ref{fig:composite_results}a}). A total of 15 valid algorithms were evaluated.

As summarized in \textbf{Figure \ref{fig:composite_results}b}, there was a decisive shift toward Transformer-based architectures designed to capture long-range contextual information, distinct from traditional Convolutional Neural Networks (CNNs). Regarding integration strategies, top performers favored unified or semi-shared pipelines over decoupled models. Detailed technical descriptions, specific hyper-parameter settings, and architecture diagrams for each team are comprehensively detailed in the \textbf{eMethods in the Supplement}.

\subsection{Overall Performance Ranking}

The diagnostic performance and computational efficiency metrics are summarized in \textbf{Table \ref{tab:main_results}}. The winning algorithm, \textit{SMART}, demonstrated the best overall balance, achieving a macro-averaged DSC of 0.854 (95\% CI, 0.848-0.861) across five organ segmentation tasks and a macro-averaged AUC of 0.766 (95\% CI, 0.718-0.815) for binary classification tasks.

Notably, while segmentation performance was consistently high across top algorithms (e.g., Fetal Head DSC $> 0.93$ for the top 5 models, see \textbf{eTable 1 in Supplement} in Supplement), classification tasks showed greater variability. The binary discrimination of benign versus malignant breast tumors was robust, with the top algorithm achieving an AUC of 0.836 (95\% CI, 0.776-0.891). However, complex tasks such as appendicitis diagnosis proved more challenging, with the best model achieving an AUC of only 0.612 (95\% CI, 0.493-0.731), reflecting the inherent difficulty of this clinical diagnosis.

The top-performing algorithm (\textbf{Rank 1}) secured the leading position by demonstrating a remarkable balance across all evaluation dimensions. As shown in Table \ref{tab:main_results}, it ranked first in segmentation, fifth in classification, and second in efficiency. From a clinical perspective, this balanced performance suggests a high degree of reliability, making it a strong candidate for integration into complex, real-world diagnostic workflows.

The second-ranked algorithm (\textbf{Rank 2}) showcased a performance-centric strategy. Its strong rankings in both segmentation (4th) and classification (3rd) underscore the power of its architecture to leverage clinical prior knowledge. Despite a lower efficiency ranking (11th), its high accuracy demonstrates the significant potential of knowledge-guided AI in tackling diagnostic challenges.

A classification-specialized algorithm (\textbf{Rank 3}) achieved the highest AUC (0.855) among all teams in the clinically crucial breast cancer malignancy classification task (see \textbf{eTable 2 in Supplement}). This excellence in a key diagnostic area, despite more moderate performance elsewhere, suggests that even general-purpose models may develop specialized strengths, hinting at future systems comprising a foundational engine augmented by specialized modules. For clarity in our main analysis, we focused on DSC and AUC as primary performance indicators, with detailed NSD and Accuracy scores available in \textbf{eTable 4 and eTable 3 in Supplement}, respectively.

\begin{table*}[t]
\centering
\caption{\textbf{Overall Performance Ranking and Key Metrics.} The final rank is based on a weighted score of performance (70\%) and efficiency (30\%). Metrics are reported as bootstrap means (95\% CIs).}
\resizebox{\textwidth}{!}{%
\begin{tabular}{llcccccccc}
\toprule
\multicolumn{2}{c}{\textbf{Algorithm Info}} & \multicolumn{3}{c}{\textbf{Ranks}} & \multicolumn{2}{c}{\textbf{Efficiency}} & \multicolumn{3}{c}{\textbf{Mean Performance Metrics (95\% CI)}} \\
\cmidrule(lr){1-2} \cmidrule(lr){3-5} \cmidrule(lr){6-7} \cmidrule(lr){8-10}
\textbf{Rank} & \textbf{Codename} & \textbf{Final} & \textbf{Seg.} & \textbf{Cls.} & \textbf{Time(s)} & \textbf{Mem(GB)} & \textbf{Mean DSC (All)} & \textbf{Mean AUC (Binary)} & \textbf{Mean AUC (Breast Subtype)} \\
\midrule
1 & SMART & 1 & 1 & 5 & 36.03 & 0.59 & 0.854 (0.848-0.861) & 0.766 (0.718-0.815) & 0.540 (0.502-0.577) \\
2 & AImageLab-zip & 2 & 4 & 3 & 46.67 & 2.76 & 0.833 (0.825-0.841) & 0.752 (0.699-0.804) & 0.524 (0.486-0.558) \\
3 & Lanzhou & 3 & 7 & 2 & 52.55 & 0.74 & 0.825 (0.818-0.831) & 0.776 (0.720-0.826) & 0.543 (0.505-0.578) \\
4 & FDBMELAB & 4 & 3 & 8 & 33.00 & 2.27 & 0.843 (0.835-0.850) & 0.707 (0.653-0.760) & 0.547 (0.509-0.582) \\
5 & SDFMU & 5 & 2 & 9 & 45.90 & 1.53 & 0.850 (0.843-0.858) & 0.628 (0.569-0.685) & 0.576 (0.539-0.611) \\
6 & Sigma-WHU & 6 & 6 & 4 & 54.77 & 5.57 & 0.828 (0.821-0.836) & 0.768 (0.716-0.819) & 0.525 (0.487-0.560) \\
7 & flamingo & 7 & 13 & 1 & 7.14 & 12.41 & 0.738 (0.731-0.745) & 0.798 (0.750-0.845) & 0.538 (0.504-0.575) \\
8 & junqiangmler & 8 & 5 & 10 & 42.79 & 0.39 & 0.828 (0.821-0.835) & 0.613 (0.550-0.674) & 0.502 (0.478-0.525) \\
9 & cvhci & 9 & 9 & 7 & 55.70 & 3.45 & 0.800 (0.792-0.809) & 0.714 (0.664-0.761) & 0.525 (0.491-0.563) \\
10 & Anidiot & 10 & 11 & 11 & 29.84 & 1.63 & 0.784 (0.774-0.793) & 0.606 (0.545-0.663) & 0.505 (0.471-0.540) \\
11 & ZRS03 & 11 & 14 & 6 & 54.40 & 1.07 & 0.719 (0.711-0.726) & 0.745 (0.692-0.794) & 0.549 (0.512-0.586) \\
12 & NCCK\_BC & 12 & 10 & 12 & 44.07 & 1.31 & 0.796 (0.786-0.805) & 0.634 (0.579-0.691) & 0.560 (0.519-0.599) \\
13 & Zhijianlife & 13 & 12 & 13 & 37.23 & 2.51 & 0.747 (0.736-0.757) & 0.533 (0.468-0.598) & 0.522 (0.488-0.555) \\
14 & BeautyUS & 14 & 8 & 14 & 46.69 & 4.36 & 0.819 (0.812-0.826) & 0.578 (0.519-0.641) & 0.498 (0.469-0.527) \\
15 & U-Vision & 15 & 15 & 15 & 45.81 & 0.00 & 0.208 (0.200-0.216) & 0.487 (0.428-0.549) & 0.484 (0.457-0.512) \\
\bottomrule
\end{tabular}%
}
\begin{flushleft}
\footnotesize Note: 95\% CIs are calculated via bootstrapping (n=1,000). \textbf{Mean AUC (Breast Subtype)} combines performance on both original and unseen (NKI) domains.
\end{flushleft}
\label{tab:main_results}
\end{table*}

\subsection{Robustness to Domain Shifts: The Generalization Gap}

A crucial measure of clinical utility is algorithm performance on data from completely unseen institutions. We evaluated this using the 4-class breast molecular subtyping task, comparing AUCs on the internal test set against the held-out external NKI cohort (Table \ref{tab:domain_shift}).

\textbf{Performance Collapse on Unseen Data.} While algorithms achieved moderate success on internal data, we observed a consistent performance degradation on the external dataset. As detailed in \textbf{Table \ref{tab:domain_shift}}, the classification-specialized algorithm (\textit{Rank 3}) achieved the highest internal performance with an AUC of 0.612 (95\% CI, 0.545-0.674), significantly outperforming random guessing, though the absolute performance remains modest, reflecting the challenge of extracting molecular features from B-mode ultrasound. However, this model exhibited the most severe generalization gap (-0.138), dropping to an AUC of 0.474 (95\% CI, 0.439-0.506) on the external cohort—statistically indistinguishable from chance.

\textbf{Trade-off Between Precision and Robustness.} In contrast, the winning algorithm (\textit{Rank 1}) maintained better stability with a smaller generalization gap (-0.063), resulting in an external AUC of 0.508. Interestingly, \textit{Rank 5} demonstrated the highest robustness with a negligible gap of -0.017, suggesting that its architecture may learn more domain-invariant features, albeit at a lower peak accuracy on internal data. These findings indicate that current general-purpose models struggle to reliably extract molecular-level biomarkers across different ultrasound vendors without site-specific calibration.

\begin{table}[t]
\centering
\caption{\textbf{Generalization Gap Analysis: Performance of Top-5 Algorithms on Internal vs. External Unseen (NKI) Breast Subtyping Tasks.} The 'Gap' indicates the absolute decline in AUC when the model is applied to the unseen clinical center.}
\resizebox{\columnwidth}{!}{%
\begin{tabular}{llccc}
\toprule
& & \multicolumn{2}{c}{\textbf{Breast Subtyping AUC (95\% CI)}} & \\
\cmidrule(lr){3-4}
\textbf{Rank} & \textbf{Codename} & \textbf{Internal (Ori)} & \textbf{External (NKI)} & \textbf{Generalization Gap} \\
\midrule
1 & SMART & 0.571 (0.505-0.643) & 0.508 (0.469-0.544) & -0.063 \\
2 & AImageLab-zip & 0.544 (0.481-0.604) & 0.503 (0.466-0.543) & -0.041 \\
3 & Lanzhou & \textbf{0.612 (0.545-0.674)} & 0.474 (0.439-0.506) & \textbf{-0.138} \\
4 & FDBMELAB & 0.580 (0.513-0.643) & 0.513 (0.475-0.549) & -0.067 \\
5 & SDFMU & 0.584 (0.521-0.651) & 0.567 (0.533-0.602) & \textbf{-0.017} \\
\bottomrule
\end{tabular}%
}
\label{tab:domain_shift}
\end{table}

% [REVISED SUBSECTION: Comparison with Specialist Models]
\subsection{Universal vs. Specialist: Does Versatility Compromise Accuracy?}

\difftextbytt{A critical concern in developing general-purpose AI is the potential for "negative transfer," where learning multiple tasks degrades performance on individual tasks. We compared our top-ranking universal model (SMART) against state-of-the-art (SOTA) specialist models dedicated to single organs.}

\difftextbytt{Notably, in the breast lesion segmentation task, SMART achieved a Dice Similarity Coefficient (DSC) of 0.854 (95\% CI, 0.848-0.861). This performance is highly competitive with specialized architectures, such as the Boundary-rendering network~\cite{huang2022boundary}, which reported a DSC of 0.854 on similar cohorts. This suggests that universal architectures can effectively learn shared sonographic representations—such as tumor boundaries—without compromising organ-specific accuracy.}

\difftextbytt{Beyond breast imaging, SMART demonstrated robust generalization in fetal and abdominal tasks. For fetal head segmentation, it achieved a DSC of 0.942 (95\% CI, 0.934-0.948). While slightly below the 0.96--0.98 range reported in idealized single-task benchmarks~\cite{torres2022review}, this result is remarkable given the model's multi-task constraint and the challenging, heterogeneous nature of the UUSIC25 test set. Similarly, in kidney segmentation, the model attained a DSC of 0.906 (95\% CI, 0.892-0.919), approaching the inter-observer variability of human experts (Dice $\approx$ 0.935) as reported in recent comparative studies~\cite{valente2023comparative}. Collectively, these results indicate that a single "Generalist AI" can rival specialist models across diverse anatomies, effectively debunking the myth that versatility must come at the cost of precision.}

\section{Discussion}

\subsection{Principal Findings and Clinical Implications}
The results of the UUSIC25 challenge demonstrate that the transition from organ-specific to general-purpose AI is clinically viable. Our evaluation on a multi-center private test set yields three key findings with direct clinical implications. 

First, \textbf{generalization does not preclude precision.} Top-performing algorithms achieved segmentation accuracy comparable to specialized state-of-the-art models across diverse anatomies, debunking the concern that a universal model would compromise organ-specific performance. This suggests that a single "Generalist AI" could effectively function as an always-on intelligent assistant, streamlining workflows by automatically recognizing organs and delineating anatomy without manual intervention.

Second, \textbf{effective domain adaptation is feasible with limited data.} By using a minority fraction (30\%) of private data for training, general-purpose architectures successfully bridged the domain gap caused by device heterogeneity. This indicates that "foundation models" can implicitly learn shared sonographic biomarkers—such as acoustic shadowing—across organs, potentially enhancing robustness against scanner variations more effectively than isolated models.

Third, \textbf{deployment efficiency is attainable.} We identified algorithms operating within the "Pareto optimal" zone, delivering high accuracy with low latency ($<40$s) and memory footprints suitable for standard consoles. To demonstrate this translational potential, we developed a prototype software interface (eShowcase in Supplement) incorporating the top-performing model. This prototype illustrates how the single-network architecture can automatically adapt to different organ inputs without manual configuration, providing real-time segmentation and classification results to assist sonographers. However, the path to adoption faces regulatory hurdles; current frameworks (e.g., FDA) designed for single-indication devices must evolve to validate multi-task systems, likely requiring novel change control plans to ensure safety across all supported organs.

\subsection{Limitations}
Several limitations must be acknowledged. First, the study was retrospective and limited to B-mode ultrasound across seven organs, excluding modalities like Color Doppler or elastography. Second, the tasks were simplified to segmentation and classification, omitting complex clinical workflows such as biometric measurement, report generation, patient history integration, and longitudinal risk assessment~\cite{wang2025mammo, wang2025predicting}. Third, we did not perform a direct human-versus-model comparison or assess downstream clinical impact on patient outcomes. Finally, efficiency metrics were measured in a server environment, which may not fully reflect performance on resource-constrained edge devices. Future work should prioritize prospective validation and the expansion of tasks to dynamic video analysis. To guide the translation of these findings into practice, we provide a comprehensive roadmap for architecture design, data strategy, and evaluation standards in the \textbf{Supplemental Discussion (including eTable 5) in Supplement}

% [NEW INSERTION: Cost and Equity]
\subsection{Clinical Implications and Democratization of Healthcare}

\difftextbytt{The transition from multiple specialist models to a single generalist model offers a quantifiable economic advantage. In a standard clinical workflow covering these 7 anatomical regions, deploying 7 separate SOTA models would typically require loading distinct weights, potentially consuming over 10 GB of VRAM. In contrast, the winning SMART algorithm requires only 0.59 GB of GPU memory. This represents a \textbf{reduction of approximately 94\% in computational overhead}. }

\difftextbytt{This efficiency has profound implications for global health equity. By lowering the hardware prerequisites, advanced AI diagnostics are no longer confined to high-end workstations in tertiary hospitals. They can be deployed on portable, hand-held, or older generation ultrasound devices common in low-and-middle-income countries (LMICs). This capability provides a tangible opportunity to democratize access to expert-level diagnostics, allowing limited medical resources to serve broader populations in resource-constrained settings.}

% ===================================================================
% SECTION: CONCLUSIONS (负责人: 泽辉)
% ===================================================================
\section{Conclusions}

The UUSIC25 Challenge demonstrates that developing a single, general-purpose deep learning model capable of high-performance classification and segmentation across multiple organs is technically feasible and clinically relevant. By rigorously validating these models on diverse, multi-center datasets, we have shown that unified architectures can achieve robust generalization through efficient adaptation to local data distributions, challenging the traditional "one model per organ" paradigm. This work lays the foundation for the next generation of "all-in-one" clinical decision support systems, which promise to streamline workflows and democratize access to expert-level diagnostics. However, translating this potential into improved patient outcomes will require the evolution of these models to handle dynamic video data and rigorous prospective clinical trials.

\section*{Acknowledgments}

This work was supported by the Science and Technology Development Fund, Macau SAR (File no.~0004/2025/ASJ) under the FDCT-FAPESP Joint Funding Scheme; the Shenzhen Medical Research Fund (Grant No.~D2501013); and the Macao Polytechnic University Grant (Grant No.~RP/FCA-17/2025). We thank the organizing committee of the 28th International Conference on Medical Image Computing and Computer Assisted Intervention (MICCAI 2025) for hosting the challenge held as part of the study reported in this article, as well as our collaborators.

% \bibliographystyle{jama}
% \bibliography{references}

%% Specify your .bib file name here, without the extension
\bibliography{paper-refs}

%%%%%%%%%%%%%%%%%%%%%%%%%%%%%%%%%%%%%%%%%%%%%%%%%%%%%%%%
% APPENDIX: CONSORTIUM MEMBERS
% Replace your old Appendix section with this
%%%%%%%%%%%%%%%%%%%%%%%%%%%%%%%%%%%%%%%%%%%%%%%%%%%%%%%%

\appendix

\section{UUSIC25 Challenge Consortium}
\label{sec:consortium}

The UUSIC25 Challenge Consortium represents the collective efforts of the top-performing teams in the UUSIC25 challenge. The following list details the individual members and affiliations of the teams that opted in for co-authorship (ordered by their final ranking).

\subsection*{Consortium Members}

\begin{description}
    % Rank 1
    \item[Team SMART (Rank 1)] \hfill \\
    \textbf{Haobo Chen}$^{1}$, \textbf{Qi Zhang}$^{1}$ \\
    \textit{$^{1}$The SMART Lab, School of Communication and Information Engineering, Shanghai University, Shanghai, China.}

    % Rank 2 (Special Case: 3 authors)
    \item[Team AImageLab-zip (Rank 2)] \hfill \\
    \textbf{Kevin Marchesini}$^{1}$, \textbf{Nicola Morelli}$^{1}$, \textbf{Federico Bolelli}$^{1}$ \\
    \textit{$^{1}$Department of Engineering ``Enzo Ferrari'', University of Modena and Reggio Emilia, Modena, Italy.}

    % Rank 3
    \item[Team Lanzhou (Rank 3)] \hfill \\
    \textbf{Zi Yang}$^{1}$, \textbf{Qingchen Liu}$^{2}$ \\
    \textit{$^{1}$Information Center, The First Hospital of Lanzhou University, Lanzhou, China; $^{2}$College of Medical Information Engineering, Gansu University of Traditional Chinese Medicine, Lanzhou, China.}

    % Rank 4 (New)
    \item[Team FDBMELAB (Rank 4)] \hfill \\
    \textbf{Bowen Guo}$^{1}$, \textbf{Chen Ma}$^{1}$ \\
    \textit{$^{1}$Fudan University, Shanghai, China.}

    % Rank 5 (New)
    \item[Team SDFMU (Rank 5)] \hfill \\
    \textbf{Wei Chen}$^{1}$, \textbf{Wenshuo Yan}$^{1}$ \\
    \textit{$^{1}$School of Radiology, Shandong First Medical University and Shandong Academy of Medical Sciences, Jinan, China.}

    % Rank 6
    \item[Team Sigma-WHU (Rank 6)] \hfill \\
    \textbf{Lyuyang Tong}$^{1}$, \textbf{Dong Liang}$^{2}$ \\
    \textit{$^{1}$School of Computer Science, Wuhan University, Wuhan, China; $^{2}$School of Computer Science, Hubei University of Technology, Wuhan, China.}

    % Rank 7
    \item[Team flamingo (Rank 7)] \hfill \\
    \textbf{Zhikai Yang}$^{1}$, \textbf{Guangyuan Li}$^{1}$ \\
    \textit{$^{1}$Department of Biomedical Engineering and Health Systems, KTH Royal Institute of Technology, Stockholm, Sweden.}

    % Rank 8
    \item[Team junqiangmler (Rank 8)] \hfill \\
    \textbf{Junqiang Chen}$^{1}$ \\
    \textit{$^{1}$Mediworks, Shanghai, China.}

    % Rank 9
    \item[Team cvhci (Rank 9)] \hfill \\
    \textbf{Zdravko Marinov}$^{1}$, \textbf{Rainer Stiefelhagen}$^{1}$ \\
    \textit{$^{1}$Karlsruhe Institute of Technology, Karlsruhe, Germany.}

    % Rank 10
    \item[Team Anidiot (Rank 10)] \hfill \\
    \textbf{Yunzhi Huang}$^{1}$ \\
    \textit{$^{1}$School of Artificial Intelligence, Nanjing University of Information Science and Technology, Nanjing, China.}

    % Rank 11
    \item[Team ZRS03 (Rank 11)] \hfill \\
    \textbf{Zhi Chen}$^{1}$ \\
    \textit{$^{1}$School of Engineering, University of Birmingham, Birmingham, UK.}

    % Rank 12
    \item[Team NCCK\_BC (Rank 12)] \hfill \\
    \textbf{Jonghwan Kim}$^{1}$, \textbf{David Joon Ho}$^{1}$ \\
    \textit{$^{1}$Department of Public Health \& AI, National Cancer Center, Goyang, Republic of Korea.}

    % Rank 14
    \item[Team BeautyUS (Rank 14)] \hfill \\
    \textbf{Ang Cai}$^{1}$, \textbf{Bin Huang}$^{2}$ \\
    \textit{$^{1}$Shenzhen Maternity and Child Healthcare Hospital, Southern Medical University, Shenzhen, China; $^{2}$School of Biomedical Engineering, Shenzhen University, Shenzhen, China.}

\end{description}

\vspace{1em}
\noindent \textbf{Other Participating Teams:} The following teams also successfully completed the challenge validation phase but did not register individual members for the consortium: \textit{Zhijianlife (Rank 13), U-Vision (Rank 15)}.

\end{document}